\documentclass{article}
\usepackage{spconf,amsmath,graphicx}
\usepackage{url}

\title{UNIVERSAL ADVERSARIAL ROBUSTNESS OF TEXTURE AND SHAPE-BIASED MODELS}

\name{Kenneth T. Co$^{\ast, \dagger}$, Luis Mu\~noz-Gonz\'alez$^{\ast}$, Leslie Kanthan$^{\dagger}$, Ben Glocker$^{\ast}$, Emil C. Lupu$^{\ast}$\thanks{Kenneth Co is sponsored in part by DataSpartan research grant DSRD201801.}
\thanks{
\textcopyright 2021 IEEE. Personal use of this material is permitted. Permission from IEEE must be obtained for all other uses, in any current or future media, including reprinting/republishing this material for advertising or promotional purposes, creating new collective works, for resale or redistribution to servers or lists, or reuse of any copyrighted component of this work in other works.}
}
\address{$^{\ast}$Department of Computing, Imperial College London, London, United Kingdom\\
$^{\dagger}$DataSpartan Research, London, United Kingdom\\
\{k.co, l.munoz, b.glocker, e.c.lupu\}@imperial.ac.uk, l.kanthan@dataspartan.com
}

\begin{document}
\maketitle
\begin{abstract}
Increasing shape-bias in deep neural networks has been shown to improve robustness to common corruptions and noise. In this paper we analyze the adversarial robustness of texture and shape-biased models to Universal Adversarial Perturbations (UAPs). We use UAPs to evaluate the robustness of DNN models with varying degrees of shape-based training. We find that shape-biased models do not markedly improve adversarial robustness, and we show that ensembles of texture and shape-biased models can improve universal adversarial robustness while maintaining strong performance.
\end{abstract}
\begin{keywords}
Universal adversarial perturbations, adversarial machine learning, deep neural networks
\end{keywords}

\section{Introduction}
Advances in computation and machine learning have enabled Deep Neural Networks (DNNs) to become the algorithm of choice for large-scale image classification \cite{krizhevsky2012imagenet, simonyan2014very, he2016deep}. To further understand the types of features that computer vision DNNs learn, \cite{brendel2019approximating, geirhos2019imagenet} we have investigated the effect of texture and shape-bias that models have on their performance. Evidence from \cite{brendel2019approximating} and \cite{geirhos2019imagenet} demonstrate that it is sufficient for models to use image textures to achieve high accuracy on ImageNet \cite{russakovsky2015imagenet}.

Geirhos et al. propose Stylized-ImageNet, a modified ImageNet dataset that requires recognizing object shapes rather than textures to discriminate images in the dataset \cite{geirhos2019imagenet}. They claim that biasing models towards shapes improves their robustness, and they show that these shape-biased models have improved robustness against common corruptions. However, to have a complete characterization of a model's robustness, it is important to also study the worst-case distortions. These worst-case distortions come in the form of \emph{adversarial perturbations}: visually imperceptible changes to inputs that result in images that fool the model \cite{biggio2013evasion, szegedy2014intriguing}. Despite their success, DNNs have been shown to be extremely sensitive to adversarial perturbations as they can be made to misclassify images with very high confidence. Adversarial attacks remain a relevant threat to shape-biased models as these attacks can greatly undermine the integrity and trust in model predictions.

Universal adversarial perturbations (UAP) are a particularly potent class of adversarial attacks where a single UAP can fool a model on a large set of input data \cite{moosavi2017universal}. From a machine learning perspective, UAPs reveal global features that models are sensitive to \cite{ilyas2019adversarial, co2021jacobian}. These global features are worth studying as they can reveal underlying features that models use for classification. From a security perspective, UAPs pose a worrying threat as they can transfer across models \cite{tramer2019adversarial, co2019procedural}, enable physically-realizable attacks on computer vision systems \cite{athalye2018synthesizing, eykholt2018robust, tu2020physically}, and can be used to facilitate efficient black-box attacks on DNNs \cite{co2019procedural}. Recent work shows that there are some improvements in adversarial robustness to per-instance attacks by increasing shape-bias \cite{shi2020informative}, however our results show that this is not the case for universal attacks. 

In this work, we evaluate texture and shape-biased models' robustness to universal attacks, propose targeted UAPs as a method for visualizing and analyzing their most vulnerable features, and finally demonstrate that ensemble voting with these models can maintain clean performance whilst improving on worst-case performance against universal attacks. Our analysis with UAPs reveals the extent to which increased shape-bias improves adversarial robustness of models to universal attacks. The following are our contributions:

\begin{figure*}[t]
\centering
\includegraphics[width=\textwidth]{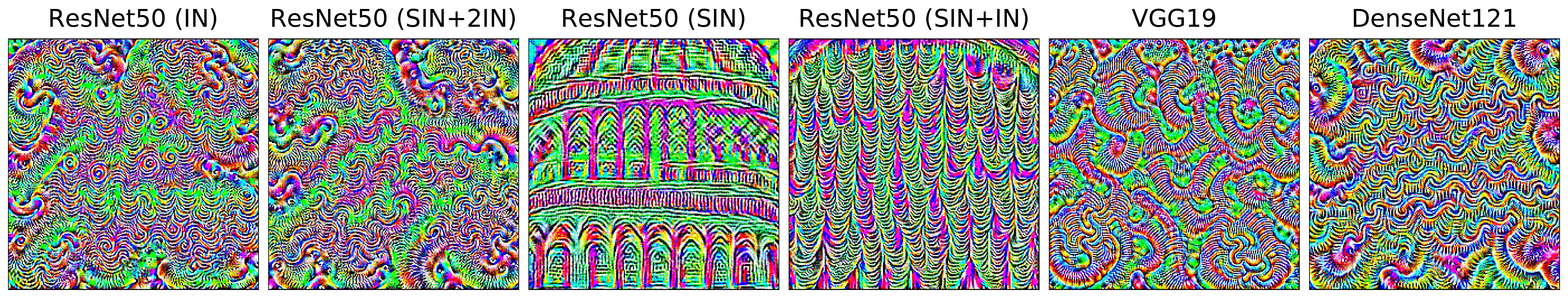}
\caption{Untargeted UAPs generated for different models, from left to right: ResNet50 \cite{he2016deep} models for various training regimes, VGG19 \cite{simonyan2014very}, and DenseNet121 \cite{huang2017densely}. Complete details for how each ResNet50 model was trained are in Section~\ref{sec:related}.}
\label{fig:perturbations}
\end{figure*}

\begin{enumerate}
    \item We show that shape-biased models do not mitigate universal attacks, but they instead shift the set of features the models are vulnerable to.
    \item We evaluate the distribution of targeted UAPs for texture and shape-biased models, and visualize the resulting targeted UAPs to show differences in the appearance of resulting perturbations.
    \item Finally, we propose ensembles of texture and shape-biased models to maintain strong performance and improve robustness to universal attacks.
\end{enumerate}

We make code available at \url{https://github.com/kenny-co/sgd-uap-torch}.
\section{Background}
\label{sec:related}
In this section, we introduce shape-biased models and universal adversarial perturbations. The Stylized-ImageNet dataset was created to train models to have higher shape-bias. To study the robustness of these models with varying degrees of shape-biased training, we first discuss how UAPs are generated and the metrics we use to measure their effectiveness.

\subsection{Shape-biased Training}
\label{sec:shape}
ImageNet is a widely used computer vision benchmark with 1,000 distinct object categories  \cite{russakovsky2015imagenet}. Geirhos et al. created a shape-biased dataset in \textit{Stylized-ImageNet}, which replaces textures of ImageNet images while retaining the global shape of the original objects \cite{geirhos2019imagenet}. This dataset requires a model to use shapes rather than textures to identify and discriminate objects. Stylized-ImageNet is generated by applying style transfer of different uninformative textures onto ImageNet images. The complete details of its generation are described in \cite{geirhos2019imagenet}.

\textbf{Models.} We use the ResNet50 \cite{he2016deep} architecture with varying degrees of training on ImageNet and Stylized-ImageNet. This model takes input images with dimensions $224 \times 224 \times 3$. We test UAPs on these four ResNet50 models, named according to the training dataset used: only ImageNet (\textbf{IN}), only Stylized-ImageNet (\textbf{SIN}), both Stylized-ImageNet and ImageNet (\textbf{SIN+IN}), and both Stylized-ImageNet and ImageNet, and then fine-tuned on ImageNet for accuracy (\textbf{SIN+2IN}). Complete training details for each model are given in \cite{geirhos2019imagenet}. From here onwards, we use IN and SIN to refer to the models rather than the datasets.

\subsection{Universal Adversarial Perturbations}
Adversarial perturbations are \textit{universal} when the same noise pattern can be successfully applied across a large fraction of the input data to fool a model \cite{moosavi2017universal}.

\textbf{Stochastic gradient descent.} We use the \textit{Stochastic Gradient Descent} (SGD) algorithm for generating UAPs. SGD is a variation of the Projected Gradient Descent (PGD) attack proposed in \cite{madry2018towards}, but optimized over batches instead of individual images. SGD was chosen as it has been shown to have the best universal evasion rates over other methods \cite{shafahi2018universal, deng2020universal}, and it is a more efficient algorithm with better convergence guarantees than the original UAP generation algorithm iterative-DeepFool \cite{moosavi2017universal}.

SGD optimizes the objective $\sum_i \mathcal{L}(x_i + \delta)$, where $\mathcal{L}$ is the model's training loss, $X_{\text{batch}} = \{x_i\}$ are batches of inputs, and $\delta \in \mathcal{P}$ are the set of considered perturbations. Gradient updates to $\delta$ are done in batches $X_{\text{batch}}$ in the direction of $-\sum_i \nabla \mathcal{L}(x_i + \delta)$. Pixel values of resulting adversarial examples $x + \delta$ are clipped to the range $[0, 255]$.

In this study, the perturbation constraints take the form of an $\ell_p$-norm. The set of perturbations can be written as $ \mathcal{P}(p, \varepsilon) = \{\delta \mid \Vert \delta \Vert_{p} \leq \varepsilon\}$ for a chosen norm $p$ and value $\varepsilon$. We choose $p = \infty$ as it is the standard in the UAP and adversarial machine learning literature. The $\ell_{\infty}$-norm constraint on $\delta$ ensures that the perturbation is small and does not greatly alter the visual appearance of the resulting image.

\textbf{Fooling rate.} We first consider the case where the goal of the UAP is to maximize the number of misclassifications. This is referred to as an untargeted UAP, as there is no specific target output. Its \textit{effectiveness} is measured by its \emph{fooling rate}, which is the proportion of inputs that are misclassified by the model when the UAP is applied. For targeted UAPs, the goal is to have as many inputs classified towards a desired target class $y_{\text{target}}$. The \emph{targeted fooling rate (tFR)} is the proportion of inputs classified as $y_{\text{target}}$ when the UAP is applied.

\section{Robustness of Shape-biased Models}
\label{sec:uap}
In this section, we compare the robustness of ResNet50 models under different training regimes by measuring the effectiveness and transferability of untargeted UAPs generated from the SGD attack. Our results show that models trained on Stylized-ImageNet are still as vulnerable to these UAPs as models trained on ImageNet.

\begin{figure*}[t]
\centering
\includegraphics[width=0.95\textwidth]{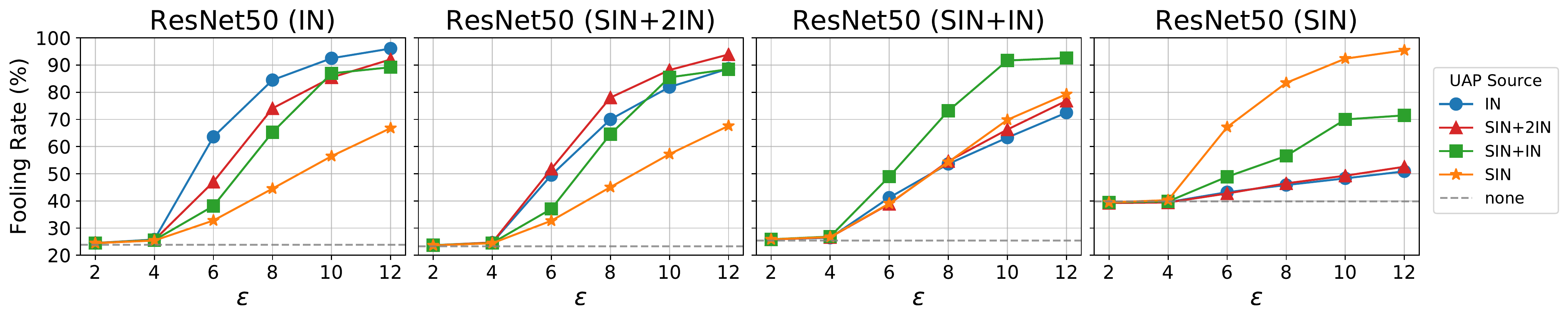}
\caption{Fooling rates (in \%) of untargeted UAPs for different perturbation values $\varepsilon$. Plot titles indicate the evaluated model.}
\label{fig:uap_eps}
\end{figure*}

\textbf{Experimental setup.} We consider the four ResNet50 models as outlined in Sec.~\ref{sec:shape}. We generate and optimize the SGD-based perturbations on each model for various $\ell_{\infty}$-norms ($2 \leq \varepsilon \leq 12$) over the entire 50,000 image ImageNet validation set. The UAP literature often focuses on $\varepsilon = 10$, so we use this as our primary benchmark, but we also provide results for other values to measure its effectiveness under different perturbation constraints. For each UAP, we evaluate its fooling rate on the model it was generated from (white-box attack) and on the three remaining models (transfer attack).

\subsection{Transferability Across Training Regimes}
White-box attacks, where the UAP is generated from the tested model, achieve high success rates. They consistently reach greater than $90\%$ fooling rate for $\varepsilon = 10$ on all models as shown in Figure~\ref{fig:uap_eps}. This shows that Stylized-ImageNet training does not necessarily improve the model robustness to universal attacks. For transfer attacks, where the UAP is generated from a model different from the evaluated one, the fooling rate consistently rises for $\varepsilon > 4$. SIN appears to be the most resilient to transfer attacks against all other models trained on the ImageNet dataset. However, this robustness to transfer attacks comes at the cost of having the highest clean error on ImageNet.

\begin{figure}[htb]
\centering
\includegraphics[width=0.8\columnwidth]{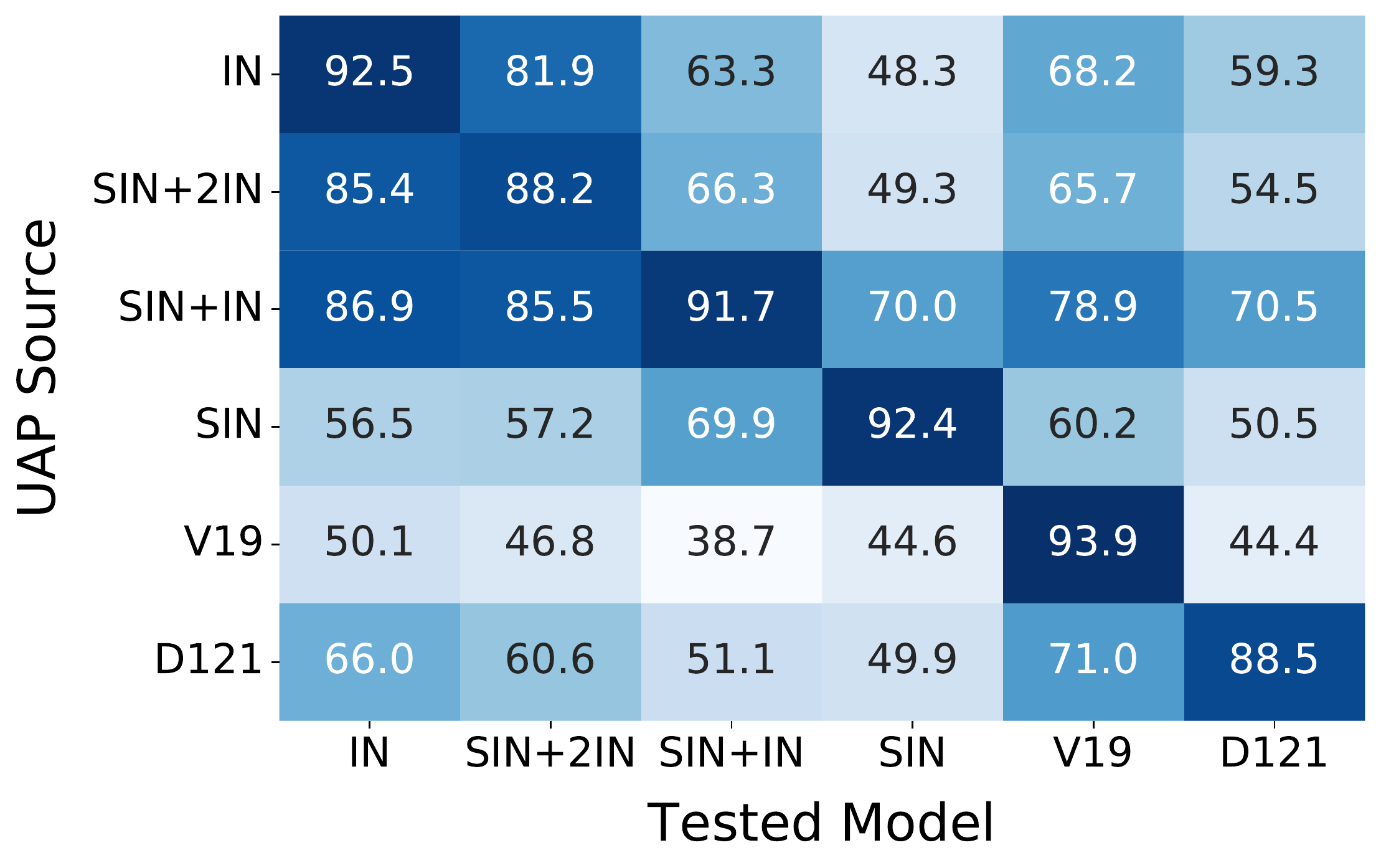}
\caption{Confusion matrix showing the fooling rates (in \%) of untargeted UAPs with $\varepsilon = 10$ when applied as transfer attacks to different models. VGG19 and DenseNet121 are labeled as V19 and D121 respectively.}
\label{fig:confusion}
\end{figure}

That white-box UAPs are still effective against SIN shows that Stylized-ImageNet training does not diminish the effects of UAPs. Instead, Stylized-ImageNet training shifts the set of features the model has learned, making it vulnerable to a different set of features from what the ImageNet-trained models have learned. Hence, what features models are most vulnerable to depends more on the data distribution of its training dataset. It is interesting to note in Figure~\ref{fig:uap_eps} that UAPs from IN and SIN+2IN have near-identical effectiveness against SIN.

\textbf{Impact of fine-tuning.} Although SIN+2IN achieves the best clean dataset accuracy, our results show that its response to UAPs is very similar to that of the model IN that was trained only on ImageNet. In Figures~\ref{fig:uap_eps} and \ref{fig:confusion}, the fooling rates of UAPs from IN and SIN+2IN are highly similar across all tested models. Additionally, there are large visual similarities between the UAPs generated for SIN+2IN and IN as seen in Figure~\ref{fig:perturbations}. These suggest that there are large similarities in the features that SIN+2IN and IN are vulnerable to, despite the additional training on Stylized-ImageNet. The additional fine-tuning on ImageNet for SIN+2IN could have resulted in the model ``overfitting'' where it no longer uses features it has learned from both datasets, and instead focuses only on features learned from ImageNet.

\subsection{Transferability Across Architectures}
We now consider the transferability of the UAPs from these ResNet50 models to other architectures: DenseNet121 \cite{huang2017densely} and VGG19 \cite{simonyan2014very}. Like ResNet50, these models take input images of dimension $224 \times 224 \times 3$ and have comparable clean dataset accuracies. Compared to ResNet50, VGG19 has fewer layers, whilst the DenseNet121 has more layers and uses dense blocks. The architectures have approximately 8, 25, and 140 million parameters for DenseNet121, ResNet50, and VGG19 respectively. We focus our analysis on attacks with $\varepsilon = 10$.

In Figure~\ref{fig:confusion}, VGG19 is noticeably more vulnerable to all UAPs, whereas DenseNet121 is more resilient to the UAPs from the other models. When focusing only on the ImageNet-trained models (IN, VGG19, DenseNet121), we see that the architectures with more layers and fewer parameters appear to be more robust. Compared to IN, SIN does not improve on the transferability of UAPs to these other architectures. However, the UAP from SIN+IN has noticeably higher transferability against all models, achieving \emph{at least} 70\% fooling rate against each model. We hypothesize that SIN+IN has learned more generalized representations because it was trained equally on both ImageNet and Stylized-ImageNet without overfitting to either, thus allowing it to have a more potent transfer attack to the other model architectures.

\begin{figure}[t]
\centering
\includegraphics[width = \columnwidth]{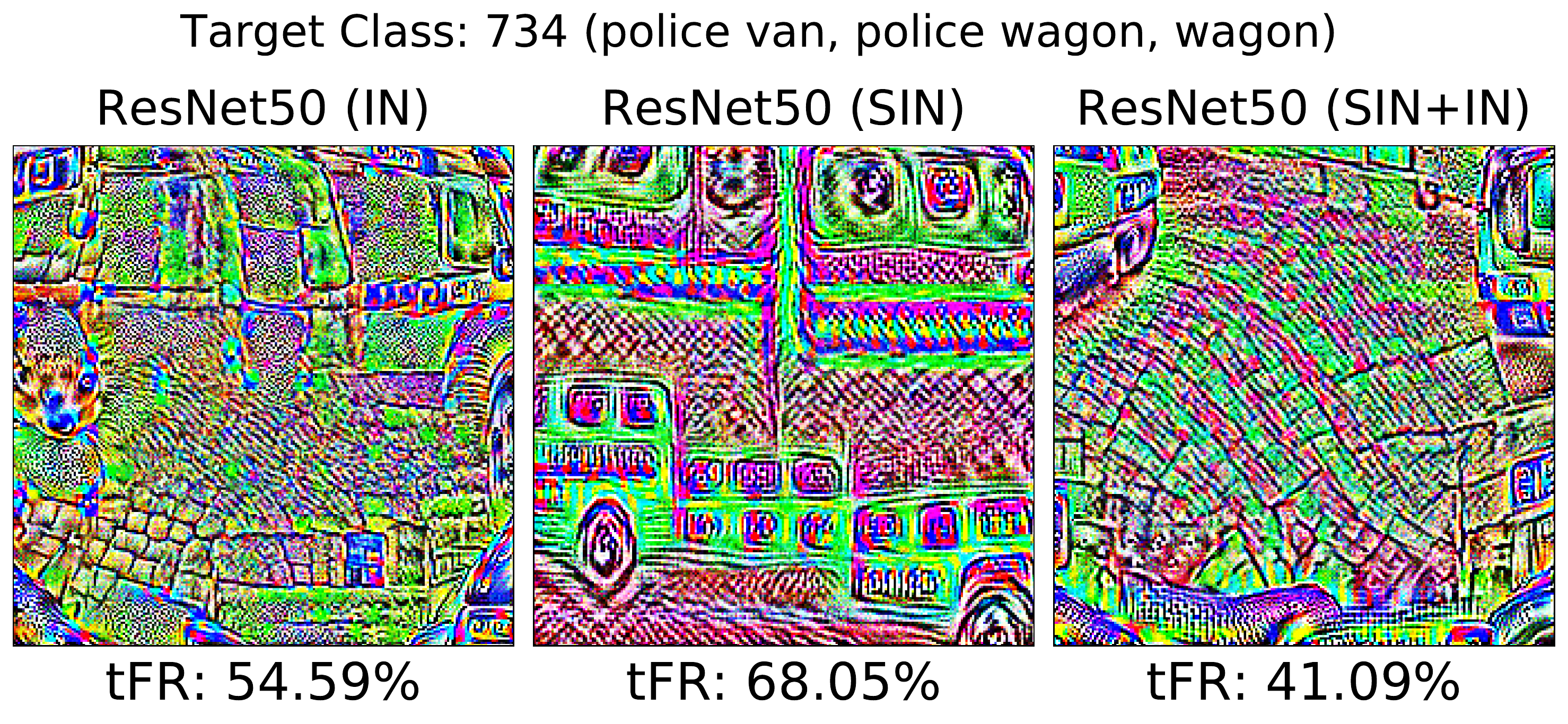}
\caption{An example of a targeted UAP and with its tFR for each model it was generated for.}
\label{fig:tuap_vis}
\end{figure}

\begin{figure}[t]
\centering
\includegraphics[width = \columnwidth]{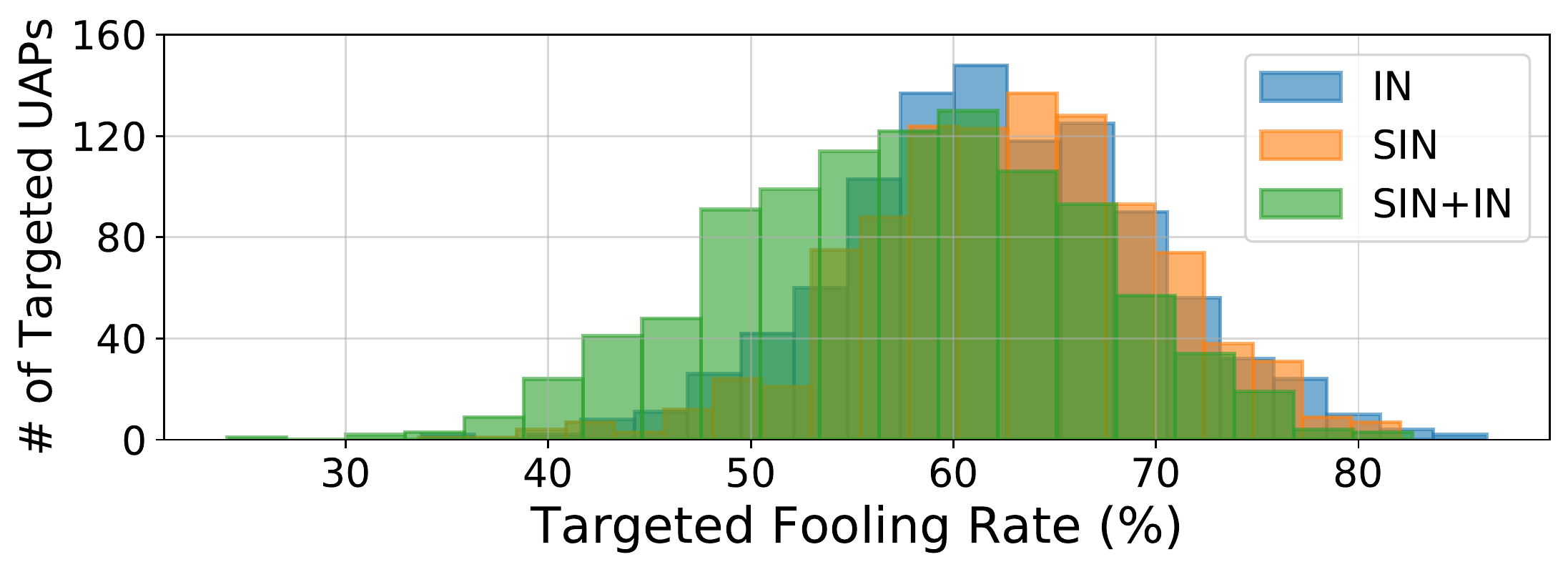}
\caption{Distribution of tFR of targeted UAPs on the ResNet50 models, with one UAP for each ImageNet class labels.}
\label{fig:tuap}
\end{figure}

\subsection{Effectiveness of Targeted Universal Attacks}
We now show how targeted UAPs can be used to visualize and analyze the sensitivity of the ResNet50 models to UAPs for each class label. We omit the analysis for SIN+2IN as its robustness to universal attacks is highly similar to that of IN.

Targeted UAPs are generated for each of the 1,000 ImageNet class labels with $\varepsilon = 10$ on each of the ResNet50 models and their effectiveness is measured by their targeted fooling rate (tFR). Figure~\ref{fig:tuap_vis} shows an example of how the different training regimes result in targeted UAPs with differing visual appearances despite targeting the same class.

Figure~\ref{fig:tuap} shows that the tFR ranges from approximately 30-80\% and the majority of the 1,000 targeted UAPs for each model are centered the around 50-70\% tFR range. What stands out is that SIN+IN appears to be slightly more robust to targeted UAPs as its tFR distribution is shifted to the left when compared to IN and SIN. Since SIN+IN blends elements from both datasets, its larger and more diverse training may have made it more robust to targeted UAPs.

\section{Mitigating Attacks with Ensembles}
\label{sec:ensemble}
Based on results from the previous section, we test an ensemble voting mechanism to improve the overall robustness to universal attacks and maintain clean performance. This would make sense as UAPs across the three models (IN, SIN, SIN+IN) do not transfer perfectly and their targeted UAPs evidence some diversity in visual appearance and effectiveness.

\textbf{Experimental setup.} To combine IN, SIN, and SIN+IN, we consider two ensemble voting schemes: a \emph{hard} voting scheme where the ensemble outputs the majority class label, and a \emph{soft} voting scheme where the ensemble outputs the class label with the highest average probability after the softmax layer. For both voting schemes, the ensemble outputs no prediction if there is no majority. To compare the ensemble scheme with the original models, we measure their accuracy on the clean dataset and the lowest accuracy achieved on any of the UAPs generated so far at $\varepsilon = 10$. This is to consider the worst-case robustness of these models to universal attacks. For IN, SIN, and SIN+IN, this would be their accuracy against their corresponding white-box untargeted UAP.

\begin{table}[htb]
\centering
\caption{Accuracy (in \%) of ResNet50 models and ensembles for the clean ImageNet validation set and on the worst-case UAP at $\epsilon = 10$. Highest model accuracy is highlighted.}
\begin{tabular}{lcc}
\\
	Model & Clean & Worst-case UAP \\
	\hline
	IN & \textbf{76.13} & 7.50\\
	SIN & 60.18 & 7.65\\
	SIN+IN & 74.59 & 8.33\\
	Ensemble (hard) & 73.24 & 17.25\\
	Ensemble (soft) & 76.02 & \textbf{20.37}\\
\end{tabular}
\label{table:ensemble}
\end{table}

\textbf{Results.} In Table~\ref{table:ensemble}, both ensembles improve on the robustness to the worst universal attack when compared to the individual models, with soft voting achieving the highest accuracy of 20.37\%, more than 10 points better than the individual models. The soft ensemble also achieves a clean accuracy of 76.02\%, which is very close to the highest clean accuracy of IN at 76.13\%. This demonstrates that the ensemble with soft voting is able to maintain clean accuracy while greatly improving adversarial robustness to universal attacks. Although the adversarial robustness of the ensemble could still be improved, this is a promising direction and demonstrates that ensemble methods can improve adversarial robustness work while maintaining clean accuracy.

\section{Conclusion}
\label{sec:conclusion}
We study texture and shape-biased models through the lens of universal adversarial robustness and show that shape-biased models are as vulnerable to universal attacks as texture-biased models. We demonstrate how UAPs can be used to better evaluate the robustness of models with differing degrees of texture and shape-biased training. Finally, we propose ensembles of texture and shape-bias models and show that it can maintain clean performance while improving on universal adversarial robustness.

\bibliographystyle{IEEEbib}
\bibliography{main}

\end{document}